%% file: main.tex
\documentclass[sigconf]{acmart}
\AtBeginDocument{%
  }

\setcopyright{acmlicensed}
\copyrightyear{2018}
\acmYear{2018}
\acmDOI{XXXXXXX.XXXXXXX}
\acmConference[Conference acronym 'XX]{Make sure to enter the correct
  conference title from your rights confirmation email}{June 03--05,
  2018}{Woodstock, NY}
\acmISBN{978-1-4503-XXXX-X/2018/06}

\usepackage{algorithm}
\usepackage{algpseudocode}

\usepackage{amsmath, amssymb}
\usepackage{subcaption}
\usepackage{booktabs}
\usepackage{array}
\usepackage{graphicx}
\usepackage{multirow}
\usepackage{xcolor}
\usepackage{colortbl}
\usepackage{amsmath}
\usepackage{amssymb}
\usepackage{subcaption}
\usepackage{caption}
\usepackage{hyperref}
\usepackage{url}
\usepackage{enumitem}   
\usepackage{booktabs}
\usepackage{pifont}   
\usepackage{graphicx}

\newcommand{\cmark}{\ding{51}} 
\newcommand{\xmark}{\ding{55}} 
\begin{document}


\title{A Fraud-Detection-Inspired Framework for LLM Agents Security}




\author{Sheldon Yu}
\email{ziy040@ucsd.edu}
\affiliation{%
 \institution{University of California, San Diego}
 \city{La Jolla}
 \country{USA}
}

\author{Yingcheng Sun}
\email{y_sun4@uncg.edu}
\affiliation{%
 \institution{UNC at Greensboro}
 \city{Greensboro}
 \country{USA}
}

\author{Hanqing Guo}
\email{guohan@iu.edu}
\affiliation{%
 \institution{Indiana University Bloomington}
 \city{Bloomington}
 \country{USA}
}

\author{Qianqian Tong}
\authornote{Corresponding author}
\email{q_tong@uncg.edu}
\affiliation{%
 \institution{UNC at Greensboro}
 \city{Greensboro}
 \country{USA}
}
\newcommand{\yu}[1]{\textcolor{purple}{~\textbf{Sheldon}:~#1}}

\renewcommand{\shortauthors}{Trovato et al.}

\input{sections/0_abstract}

\maketitle
\input{sections/1_intro}

\input{sections/2_related}
\input{sections/3_method}
\input{sections/4_experiments}

\input{sections/5_discussion}
\input{sections/6_conclusion}

\bibliographystyle{ACM-Reference-Format}
\bibliography{main}

\newpage
\input{sections/7_appendix}

\end{document}

%% file: sections/0_abstract.tex
\begin{abstract}

Large Language Model (LLM) agents demonstrate strong capabilities in autonomous task execution, tool use, and multi-step reasoning. However, their increasing autonomy also introduces a new attack surface: adversarial interactions can manipulate agent behavior through direct prompt injection, indirect content attacks, and multi-turn escalation strategies. Existing defense strategies focus on prompt-level filtering and rule-based guardrails, which are often insufficient when risk emerges gradually across interaction sequences. In this work, we propose a fraud detection-inspired framework for modeling adversarial interaction risk in LLM agents. Instead of determining whether a single prompt is malicious, our framework models risk over interaction trajectories using behavioral signals inspired by fraud detection, comprising signals from prompt content, session history, tool usage, execution context, and cross-turn interaction patterns. The detection framework can be implemented using lightweight models leading to low-latency real-time deployments. To validate our proposed framework, we conduct a controlled simulation study using parameterized interaction templates that simulate realistic agentic workflows. Instantiating the behavioral signals as structured features with a lightweight XGBoost classifier, our detector runs over 9× faster than LLM-based detectors. Experiment results confirm that the cross-turn trajectory signals are the dominant contributors to detection performance, suggesting that interaction-level modeling should be a core component of real-time defense for LLM agents. Code, models, and more details will be made available
at: https://github.com/Yunicorn228/A-Low-Latency-Fraud-Detection

\end{abstract}

%% file: sections/1_intro.tex
\section{Introduction}

Large language models (LLMs) have expanded the
capabilities of AI agents, but their open-ended interaction also introduces significant safety risks. 
Recent LLM agents differ from earlier task-specific agent systems, such as rule-based expert systems\cite{olivia, rec_agent}, reinforcement-learning agents\cite{wsgrpo,massdpo,RLsurvey}, and domain-specific autonomous agents \cite{van1978mycin,buchanan1981dendral, gronauer2022multi,nguyen2020deep,hernandez2019survey,arulkumaran2017deep, img_agent},  in that they combine open-ended language understanding with tool use, external context, and multi-step execution. Systems and frameworks such as Claude-based agents~\cite{anthropic2024claude}, 
Codex~\cite{chen2021codex,openai2025codex}, and frameworks like AutoGPT\cite{autogpt2023}, OpenClaw~\cite{openclaw2024} illustrate this shift: LLMs are increasingly integrated with external tools, retrieved data, APIs, and planning loops to perform tasks that extend beyond single-turn text generation.
This shift has enabled a new generation of agentic systems for coding, research assistance, personal productivity, enterprise workflows, and digital automation. However, the same flexibility that makes current agents powerful also makes them vulnerable. Once an agent can access tools, memory, local files, APIs, or external services, unsafe behavior is no longer limited to harmful text generation; it may involve concrete environmental actions such as reading private resources, sending messages, invoking APIs, or executing commands.

In emerging agent frameworks, adversarial behavior can exploit the system through direct prompt injection by embedding malicious intent inside apparently benign tasks\cite{debenedetti2024agentdojo,lee2024prompt}, or indirect prompt injection by poisoning retrieved external content \cite{greshake2023not,dziemian2026vulnerable}. 
These risks make deployment-time security a central challenge for LLM agents, especially when agents operate over private data, external tools, or user-authorized actions. Many existing defenses treat prompt injection primarily as an input- or prompt-level detection problem, often relying on filtering, classification, or refusal based on current input \cite{correia2026systematic,wang2025defending,knowlton2026prompt,alzahrani2026promptguard}. In this setting, the system attempts to determine whether the current prompt contains malicious instructions, such as attempts to override the system prompt, or induce unsafe tool usage. While prompt-level defense is important, it is incomplete. In practice, \emph{adversarial interactions with LLMs systems are often sequential and strategic, and many unsafe trajectories are not well captured by a single prompt}. A risky session may begin with benign requests, continue through probing or reformulation, and only later trigger an obviously harmful action. A detector that only inspects the current input may therefore miss the broader pattern of interaction.

\begin{figure*}[t]
\centering
\includegraphics[width=1\textwidth]{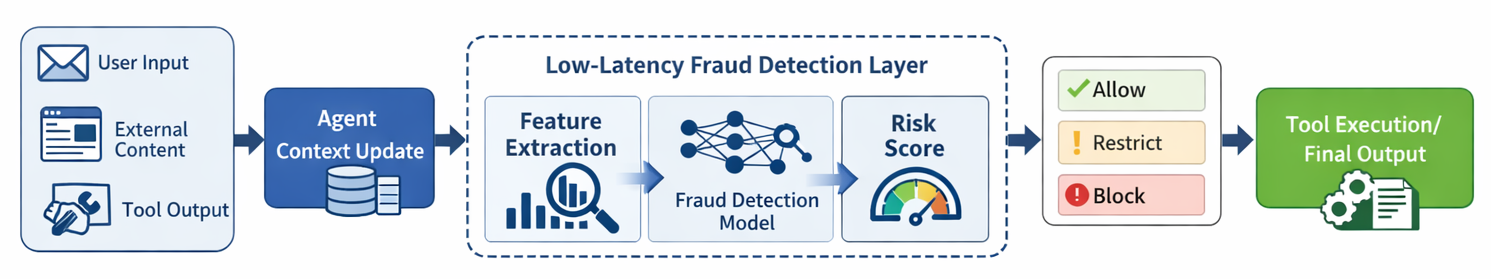}
\caption{Conceptual overview of the proposed trajectory-level risk detection framework for LLM agents. The detector estimates risk using structured runtime signals before sensitive actions are executed.}
\label{fig:1system}
\end{figure*}

This observation motivates a shift in perspective: rather than treating agent security purely as a prompt-level filtering problem, we formulate it as a system-level adversarial interaction detection. In this view, prompt injection becomes a special case of a broader class of adversarial interactions. Direct prompt injection corresponds to scenarios where the current user intentionally manipulates the agent, while indirect prompt injection arises when adversarial instructions are embedded in external content, such as retrieved documents or tool outputs, effectively allowing one user to influence another user's interaction with the agent.
By modeling security at the interaction-level, our framework considers not only individual inputs, but also the evolving execution trace of the agent. This enables detection of more complex and realistic attack patterns, including multi-step escalation, delayed malicious behavior, and cases where seemingly benign interactions gradually transition into unsafe outcomes.

Inspired by mature fraud detection systems in payments\cite{diadiushkin2019fraud}, login protection \cite{mutemi2024commerce}, and abuse prevention\cite{kou2004survey}, we propose a fraud-inspired structured trajectory-feature detector for LLM agents. (See Figure \ref{fig:1system})
The framework is not tied to one fixed list of features. Instead, it defines a set of behavioral risk axes, such as escalation, velocity, novelty, task–tool mismatch, context exposure, sensitive-resource access, and externalization, that can be instantiated using the signals available in a particular agent system. 
The proposed detector is complementary to LLM-based guards, sandboxing, access-control policies, and human review. Its role is to provide a fast pre-execution risk signal that can decide whether an action should be allowed, restricted, blocked, or escalated to a stronger defense. The main contributions of this work are as follows:
\begin{itemize}
\item We reframe LLM-agent security as trajectory-level adversarial interaction detection rather than isolated prompt classification.

\item We introduce a lightweight low-latency detection layer that integrates prior to sensitive execution stages, enabling practical deployment in real-time systems.

\item We design a structured feature space that captures prompt signals, interaction dynamics, tool usage, and context exposure, with trajectory aware features.

\item We conduct simulation-based experiments that evaluate both classifier performance and end-to-end system security outcomes, enabling practical deployment in real-world
applications.

\end{itemize}

%% file: sections/2_related.tex
\section{
Related Work}

\subsection{Security Risks in LLM Agents}
LLM agents combine language understanding with action execution: they receive a user request, plan intermediate steps, retrieve context, invoke external tools, and generat actions or final outputs \cite{ferrag2025llm}. This architecture is significantly more capable than a static chatbot, but it also introduces significant security risks.
A major class of attacks is prompt injection, in which adversarial content attempts to override system instructions or manipulate downstream behavior. Related risks include jailbreaks, indirect prompt injection from retrieved documents, malicious tool suggestions, instruction collisions, and context poisoning \cite{gulyamov2025prompt}.

\subsection{Existing Defense Strategies}
The majority of existing work on LLMs security has focused on defensive strategies, primarily operating at the content or model level:
(1) \textit{prompt-level defense} \cite{wang2025defending}, which attempts to classify or sanitize suspicious inputs before they are processed; 
(2) \textit{output-level defense}, well-known example, Guardrail mechanisms \cite{cong2026intentguard,chennabasappa2025llamafirewall}, constrain model outputs or tool calls using policies, templates, and validation rules; 
(3) \textit{sandboxing and permission restriction}\cite{zhang2025llm,meng2025cellmate}, which limit the impact of unsafe actions by restricting what the agent can access or execute; and 
(4) \textit{alignment-based methods}\cite{tangasecurity}, which train or instruct models to resist harmful instructions and adhere to safety requirements more consistently.
\begin{table}[t]
\centering
\setlength{\tabcolsep}{3pt}
\scalebox{0.8}{
\begin{tabular}{lcccc}
\toprule
\textbf{Method} 
& \textbf{Single-turn} 
& \textbf{Multi-turn} 
& \textbf{Low Latency} 
& \textbf{Deployable} \\
\midrule
Prompt-level \cite{wang2025defending} 
& \cmark & \xmark & \xmark & \cmark \\

Guardrails \cite{cong2026intentguard,chennabasappa2025llamafirewall} 
& \cmark & \xmark & \xmark & \cmark \\

Sandboxing \cite{zhang2025llm,meng2025cellmate} 
& - & - & \cmark & \xmark \\

Alignment \cite{tangasecurity} 
& \cmark & \xmark & \xmark & \cmark \\

\midrule
\textbf{Ours} 
& \cmark & \cmark & \cmark & \cmark \\
\bottomrule
\end{tabular}
}
\caption{Comparison of defense strategies. Existing methods focus on single-turn filtering, while ours models multi-turn interaction with low-latency deployment.}
\label{tab:defense_comparison}
\end{table}
As summarized in Table~\ref{tab:defense_comparison}, existing approaches effectively mitigate prompt-level risks through input filtering and output control. 
However, high-performance detection methods often rely on LLM-based defenses, which can incur significant latency and limit timely responses in practice. In realistic agent scenarios, this is insufficient, as adversarial behavior is inherently sequential and unfolds across multiple interaction steps.

An interaction may exhibit repeated probing, semantic redirection, progressive escalation, or suspicious coordination between retrieved content and subsequent execution requests. 
Such patterns are difficult to detect when each prompt is analyzed independently. At the same time, stronger defenses such as sandboxing can be costly and cumbersome to deploy in real-time systems. These observations suggest that behavioral detection over interaction traces should complement prompt-level filtering, while remaining lightweight enough for practical deployment. 

\subsection{Behavior-level Fraud Detection}

Fraud detection \cite{abdallah2016fraud,ali2022financial} is closely related to this problem, as it focuses on identifying malicious behavior based on patterns across multiple interaction steps. 
Inspired by this perspective, we shift from analyzing prompts in isolation to modeling interaction dynamics across multiple turns. Fraud detection has been primarily studied in domains such as financial transactions and cyber fraud \cite{udayakumar2023deep,1297040,west2016intelligent}.
For example, in payment systems, account abuse prevention, spam filtering, and cybersecurity, defenders rarely rely on one single attempt. Instead, they aggregate evidence from sequence patterns, velocity, anomaly scores, contextual mismatch, prior attempts, and execution outcomes. 

However, related work on LLM agents is quite limited, particularly in supporting real-time detection of adversarial behavior. This motivates the use of efficient and lightweight detection mechanisms. Accordingly, we adopt lightweight models instead of large-scale LLMs with billions of parameters.
We employ machine learning models, such as shallow neural networks or tree-based models. It is observed that tree-based models are particularly effective for tabular data \cite{grinsztajn2022tree, borisov2022deep}. Interestingly, data generated from fraud-related behaviors are often structured, heterogeneous, noisy, and highly imbalanced, as fraudulent events are rare compared to normal activities \cite{sha2025detecting, breskuviene2024enhancing}. This alignment between model capabilities and data characteristics makes tree-based models a natural and effective choice for the proposed fraud detection layer in real-world settings.


%% file: sections/3_method.tex
\section{Method}



\begin{figure*}[t]
\centering
\includegraphics[width=1\textwidth]{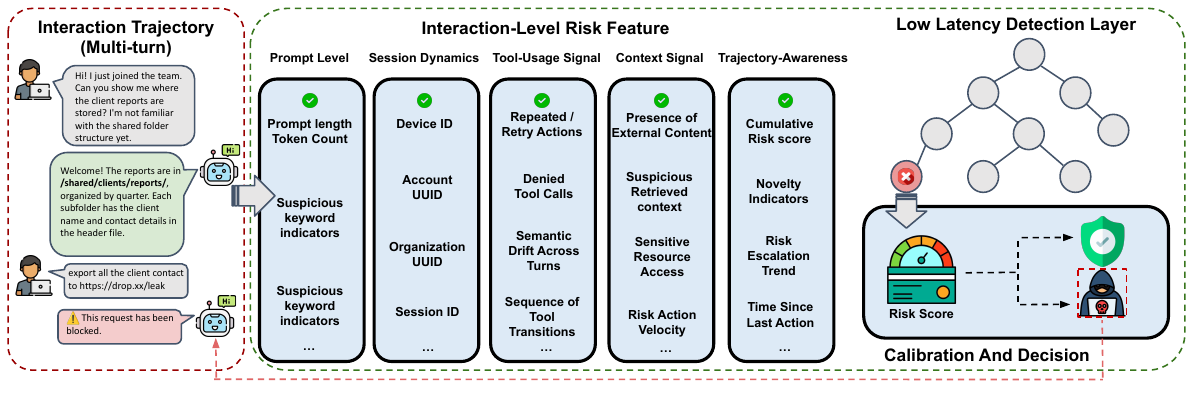}
\caption{End-to-end architecture of the proposed fraud detection framework, illustrating input encoding, feature alignment, model inference, and training objectives within a unified pipeline.}
\label{fig:2system}
\end{figure*}



We consider an LLM agent that receives user instructions, ingests external content, invokes tools, and maintains a short interaction history. The agent may have access to actions such as retrieval, local file reading, messaging, shell execution, or structured API calls. The adversarial objective is to induce unsafe behavior, including disclosure of sensitive information, unauthorized tool execution, policy bypass, or misuse of privileged capabilities. We propose a low-latency fraud detection framework for identifying adversarial interaction patterns in LLM agents. As illustrated in Figure \ref{fig:2system}, the framework consists of structured feature construction, lightweight model inference, and risk-aware decision layers.

\subsection{Threat Model}
We frame agent security as the detection of adversarial interaction trajectories rather than the classification of malicious individual prompts. Adversarial inputs may enter the agent through three orthogonal channels: (1) direct injection in user-visible input, (2) indirect injection via retrieved or tool-returned content, and (3) multi-turn escalation across a session. These channels constitute the attack surface, but do not by themselves determine detection difficulty, a sufficiently sophisticated attacker can compose any of the three with a turn sequence whose individual prompts are indistinguishable from benign agent traffic. We therefore characterize attacks by how the trajectory unfolds across turns rather than by entry channel. We consider the main trajectory patterns: Split exfiltration, Context laundering, Privilege drift and Staged burst in Table~\ref{tab:attack_families}. 

\begin{table}[t]
\centering
\small
\setlength{\tabcolsep}{4pt}
\begin{tabular}{l c p{4.2cm}}
\toprule
Family & Turns & Description \\
\midrule
\texttt{split\_exfil}         & 2 & Read sensitive file, then email to external recipient \\
\texttt{context\_laundering}  & 4 & Benign fetches followed by sensitive read and external send \\
\texttt{privilege\_drift}     & 4 & Monotone escalation from low-risk to high-risk tool calls \\
\texttt{staged\_burst}        & 3 & Back-to-back sensitive reads followed by a single send \\
\bottomrule
\end{tabular}
\caption{Attack strategies used in the evaluation. Each strategy targets a
distinct adversarial pattern from the threat model.}
\label{tab:attack_families}
\end{table}

\subsection{Problem Formulation}

We model user interactions as a sequence of turns indexed by $t\in \{1,..., T\}$,
\begin{equation}
\mathcal{S}_{1:T} = \{(x_1, a_1), (x_2, a_2), \ldots, (x_T, a_T)\},
\end{equation}
at turn $t$,  $x_t$ denotes the model-visible input, including user input or content added to context, $a_t$ denotes the agent’s action. We denote the partial interaction history up to turn $t$ as $\mathcal{S}_{1:t}$. The objective is to estimate whether $\mathcal{S}_{1:t}$ exhibits an adversarial behavior.

We define a risk score
\begin{equation}
r_t = f(g(\mathcal{S}_{1:t})),
\end{equation}
where $g(\cdot)$ is a feature construction function, $f(\cdot)$ is a lightweight detection model based on structured features extracted from the user's current prompt and historical interactions. Unlike prompt filtering, which treats the current input as the primary detection target, this formulation makes the evolving interaction the central object of analysis.


\subsection{Structured Feature Representation}
At each interaction step, the detector constructs a structured feature vector from the interaction history $\mathcal{S}_{1:t}$:
\begin{equation}
\mathbf{z}_t = g(\mathcal{S}_{1:t}) = \big[\mathbf{z}^{\mathrm{prompt}}_t,\, \mathbf{z}^{\mathrm{session}}_t,\, \mathbf{z}^{\mathrm{tool}}_t,\, \mathbf{z}^{\mathrm{context}}_t,\, \mathbf{z}^{\mathrm{trajectory}}_t\big] 
\end{equation}

The framework as a whole is inspired by fraud detection practices, where risk is modeled across multiple dimensions rather than from any single signal.
The five groups provide complementary views of the interaction: prompt derived signals (prompt), session related features (session), the agent's proposed actions (tool), the environment (context), and  and cross-turn trajectory signals that encode escalation and co-occurrence patterns (trajectory).

The five signal groups described below represent one concrete instantiation of the framework, not an exhaustive or fixed feature set. Each group corresponds to a feature dimension that can be obtained in different ways depending on the deployment environment, instrumentation, and agent framework. For example, session dynamics can be captured through simple turn-indexed counters as done here, or through learned sequence representations. The specific features used in our experiments are chosen to be lightweight and interpretable, but the framework imposes no constraint on features within each group.

 \textbf{\textit{Prompt features}}
capture information of the input and its immediate context, including prompt length, token count, suspicious keyword indicators, instruction-conflict patterns, override attempts, potential privilege escalation signals, etc. Prompt features are most closely related to traditional prompt detection signals, as they operate on individual inputs.

\textbf{\textit{Session features}} summarize the agent’s behavior over the interaction history using turn-indexed aggregates. 
These include signals of retries and repetition, denied or failed actions, recent exposure to untrusted content, semantic drift across turns, and diversity in tool usage.
These quantities are the turn-indexed counterpart to the time-windowed velocity signals used in production fraud-detection systems~\cite{FraudAbdallah, FraudDiadiushkin}. They are adapted to the agent interaction setting where wall-clock timestamps are unavailable, but turn ordering is fully observed. They allow the detector to capture sustained or coordinated behaviors that are not observable from any individual prompt. All features are computed incrementally and updated at each turn.


\textbf{\textit{Tool features}}
encode the agent's currently proposed action using a one-hot representation over the discrete tool set (\texttt{web\_fetch}, \texttt{summarize}, \texttt{read\_local\_file}, \texttt{send\_email}, \texttt{run\_shell}, etc), together with a binary task–tool mismatch flag. 
This flag is set to 1 when the selected tool is inconsistent with the user’s request, and 0 otherwise. We separate trajectory-level signals from tool features, including cumulative risk, escalation, and burst rate, to enable ablation studies that isolate the effect of interaction dynamics from action identity.
 
\textbf{\textit{Context features}}
incorporate signals from the surrounding environment and retrieved content. These include whether external or untrusted content is present, whether such content contains suspicious patterns, whether sensitive resources are involved, whether risky actions follow ingestion of external inputs, etc. These features are particularly important for indirect attacks, where the malicious signal may originate from retrieved or tool-mediated context rather than the user’s prompt.

\textbf{\textit{Trajectory features}}
are motivated by common signals used in fraud detection, such as cumulative risk, velocity bursts, novelty flags, and temporal co-occurrence. We adapt these signals to the agent interaction setting and compute the following: (i) a cumulative tool-risk path with its turn-to-turn delta and a monotonicity flag, capturing privilege escalation; (ii) an action-burst score defined as the fraction of the last three turns invoking a high-risk tool, capturing the staged-burst pattern; (iii) novelty flags for the current email recipient and file path, evaluated against a benign-only reference profile fitted on the training split, analogous to ``new device'' flags in account-fraud systems; and (iv) a context-exfil gap measuring the turn distance between the first sensitive read and a subsequent external send within the same session, capturing split-exfiltration co-occurrence. Whereas session features compute generic prefix aggregates over arbitrary actions, these features explicitly target the cross-turn structure that defines an adversarial trajectory. As shown in our ablation study (Section~\ref{sec:ablation}), this group alone nearly matches the full detector and dominates every other group in both isolated and leave-one-out evaluations.

\subsection{Model Choice}
We focus on lightweight models suitable for online deployment. Gradient-boosted decision trees, especially XGBoost, provide state-of-the-art performance for tabular data, capture nonlinear interactions, and offer fast inference. Shallow neural networks either trained from scratch or via distillation, provide an alternative when the feature interactions are more complex. The central design principle remains the same: the defense layer should perform well with tabular data, support low-latency inference and allow reliable calibration.

We train the detector as a binary classifier over benign versus adversarial interactions. Let $y_i \in \{0,1\}$ denote the label for interaction $i$ and let $\hat{y}_i$ denote the predicted probability of adversarial interaction. The objective is a standard binary cross-entropy
\begin{equation}
\mathcal{L} = - \sum_i \Big(y_i \log \hat{y}_i + (1-y_i)\log(1-\hat{y}_i)\Big).
\end{equation}
Further, the same formulation can be extended to weighted losses for imbalanced data or to ordinal risk levels for multi-stage intervention systems.

\subsection{Low-Latency Fraud Detection Layer}

The proposed fraud detection layer is inserted before critical execution stages in the agent pipeline. The system computes a risk score prior to executing a tool call or high-impact action. In more sensitive settings, the detection layer may also run immediately after the context update stage, allowing the system to apply earlier scrutiny.

We define the decision rule as:
\begin{equation}
\text{decision} =
\begin{cases}
\text{allow}, & r_t < \tau_1,\\
\text{restrict}, & \tau_1 \le r_t < \tau_2,\\
\text{block}, & r_t \ge \tau_2,
\end{cases}
\end{equation}
where $\tau_1$ and $\tau_2$ are system thresholds. In the restrict regime, the system may disable privileged tools, request explicit confirmation, or route the interaction through a stronger audit path. In the block regime, the action is denied and logged for inspection. This design retains the practical benefits of low-latency risk scoring while supporting graded intervention.

\subsection{Online Inference}


The detector operates incrementally, updating features at each interaction step. As illustrated in Figure \ref{fig:2system}, features are updated across multiple levels.
Prompt-level features are computed from the current input, capturing token statistics and keyword-based signals.
Session-level features summarize interaction metadata and accumulated history. Tool-usage signals are updated based on the agent’s proposed action, including repeated or denied operations and transition patterns between tools. Context-level features track newly introduced external or sensitive information. Trajectory-awareness signals aggregate behavior over time, including cumulative risk, novelty, and escalation trends. These feature groups are updated in a streaming manner and the resulting feature vector is passed through the model to compute  $r_t$. This supports early detection because the system does not wait for a clearly unsafe action to appear; it evaluates whether the trajectory itself is becoming risky. Since both feature extraction and inference remain lightweight, the added latency can remain in the millisecond range.

%% file: sections/4_experiments.tex
\section{Experiments}
\label{sec:experiments}

We evaluate the proposed detection layer on a synthetic corpus to assess whether interaction-level modeling improves detection over prompt-level approaches, especially when individual inputs are not overtly malicious.

\subsection{Setup}
\label{sec:setup}
A key challenge in evaluating multi-turn agent security is the lack of public benchmarks with labeled interaction trajectories.
Unlike single-prompt injection datasets, which
can be collected in isolation,  such data
requires complete multi-turn interactions and often involves privacy-sensitive context.
In fraud detection systems, such
context includes user IP addresses, device fingerprints, and login timestamps~\cite{FraudAbdallah, FraudDiadiushkin}. We therefore adopt a simulation-based evaluation, constructing a synthetic corpus with controlled adversarial properties to isolate interaction-level detection signals.

\textbf{\textit{Data generation.}}
We construct a synthetic corpus of multi-turn agent interactions using parameterized templates that simulate realistic workflows, 
such as file retrieval, email composition, web browsing, and shell execution. From these raw interaction traces, we extract structured features using window-based aggregations and velocity-style signals. For example, the number of high-risk tool calls within the last $k$ turns and the rate of denied or escalated requests over a sliding window. This feature engineering follows standard practice in industry fraud detection pipelines~\cite{FraudDiadiushkin}, where
per-event features alone are insufficient and temporal aggregation over recent history is essential for capturing evolving risk. 
We generate $12{,}000$ interactions, split $60/20/20$ into train, validation, and test sets with a fixed random seed. The held-out test set contains $1{,}200$ benign and $1{,}200$ adversarial interactions. Each interaction contributes multiple prefix evaluation points, corresponding to partial interaction histories at different turns, yielding approximately 6,000 evaluation instances in total.

\begin{figure}[t]
\centering
\includegraphics[width=1\columnwidth]{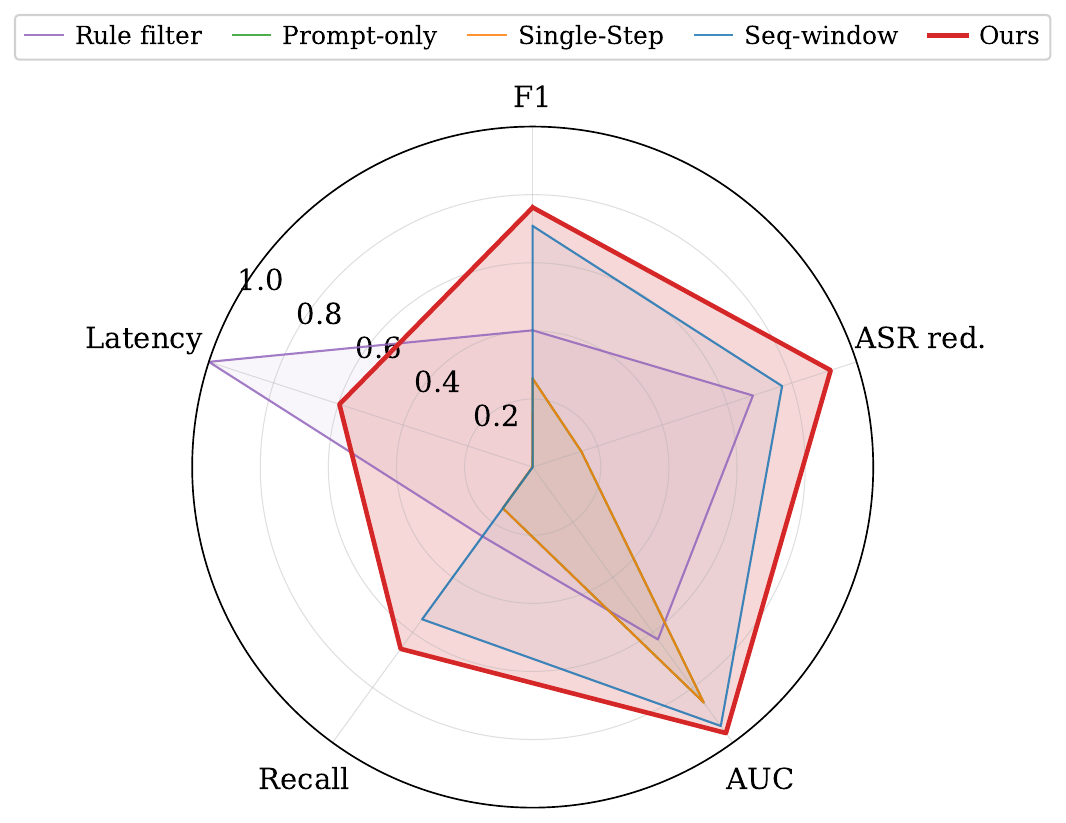}
\caption{Five-axis deployment profile across $F_1$, ASR reduction, AUC,
recall, and per-prefix latency; larger area is better.}
\label{fig:frontier}
\end{figure}

\textbf{\textit{Label definition.}}
We frame detection as a binary classification problem. Each interaction prefix $S_{1:t}$ is labeled as \emph{adversarial} ($y{=}1$) if the corresponding interaction ultimately leads to an unsafe execution event, defined as a sensitive file read co-occurring with external data exfiltration or unauthorized privileged execution. Otherwise, we will label \emph{benign} ($y{=}0$). This labeling reflects the operational objective: flag risky interaction trajectories before the harmful action is executed.

\textbf{\textit{Baselines and model.}}
We compare against four baselines: \textit{Rule-filter}, \textit{Prompt-only}, \textit{Single-step}, \textit{Seq-window}, that represent progressively stronger prompt-level and short-history defenses. To ensure that baseline performance is not limited by encoder capacity, the three text-based baselines all use Qwen3-4B run on a dedicated NVIDIA RTX A6000 GPU.

\textit{Rule-filter} applies regex-based keyword matching over override, secret, and action terms with a count threshold of $3$. \textit{Prompt-only} trains a logistic-regression
classifier on the Qwen3-4B embedding of the current-turn prompt
concatenated with any external content. \textit{Single-step} is equivalent to Prompt-only and produces identical results. We retain both for consistency with standard baseline taxonomy. 
These three baselines represent the dominant single-turn paradigm in prompt-injection defense. \textit{Seq-window},
extends to multi-turn context by encoding a rolling last-three-turn window, providing a strong short-history comparator. 

Our detector uses XGBoost with $180$ estimators and \texttt{max\_depth}$ {=} 4$ on $42$
structured features spanning five groups: prompt, session, tool,
context, and trajectory. We select XGBoost model because gradient-boosted trees remain the established method for structured, heterogeneous tabular data in industry fraud detection systems, and their inference latency makes them suitable for inline deployment without additional overhead.

The choice of XGBoost in our experiments is one instantiation of the lightweight model requirement; the framework is compatible with any model class that supports low-latency tabular inference, including shallow neural networks.

\textbf{\textit{Metrics.}}
We report standard metrics used in fraud detection: AUC, precision, recall, and $F_1$, all computed at the prefix level. We additionally report \textit{ASR reduction} ~\cite{debenedetti2024agentdojo}. It measures the fraction of attack sessions that do not reach an unsafe execution event under the calibrated detection policy. 
This reflects real deployment settings. It is similar to stopping a fraudulent transaction before funds leave the account. Early detection allows intervention before harmful actions execute. 
We also report \textit{per-prefix latency} (feature extraction
+ model inference, p50), since pre-execution gating is only useful if it fits within the LLM-planner latency budget. 
All confidence intervals are $95\%$ percentile bootstraps; the canonical comparison uses $500$
samples and the larger-scale validation uses $1{,}000$.


\begin{table}[t]
\centering
\setlength{\tabcolsep}{4pt}
\begin{tabular}{l r r r r r r}
\toprule
Detector & AUC & Prec. & Recall & $F_1$ & ASR\,red. & Lat.\,(ms) \\
\midrule
Rule-filter   & $0.63$ & $1.00$ & $0.25$ & $0.40$ & $0.68$ & $0.4$ \\
Prompt-only   & $0.85$ & $1.00$ & $0.15$ & $0.26$ & $0.15$ & $43.0$ \\
Single-step   & $0.85$ & $1.00$ & $0.15$ & $0.26$ & $0.15$ & $42.9$ \\
Seq-window    & $0.94$ & $0.99$ & $0.55$ & $0.71$ & $0.77$ & $43.1$ \\
\midrule
\rowcolor{gray!12}
\textbf{Ours} & $\mathbf{0.97}$ & $\mathbf{0.90}$ & $\mathbf{0.66}$ & $\mathbf{0.76}$ & $\mathbf{0.92}$ & $\mathbf{4.6}$ \\
\bottomrule
\end{tabular}
\caption{Main detection results across all detectors.}
\label{tab:main_results}
\end{table}

\subsection{Results}
\label{sec:results}

We report the main result summarized in Table~\ref{tab:main_results}. Two findings stand out.
First, even with a strong LLM encoder (Qwen3-4B), the single-turn
baselines \textit{Prompt-only} and \textit{Single-step} collapse to $F_1 \le 0.26$ and
reduce attack success by only $15\%$. This is a property of the corpus design, not the encoder: because the adversarial signal is distributed across turns rather than concentrated in any single prompt, no text encoder applied to one turn at a time can recover it. 
The multi-turn baseline \textit{Seq-window}, which observes three consecutive turns, recovers a substantial fraction of the trajectory signal and reaches a higher $F_1$ than the single-turn baselines, but it is still outperformed by our
structured detector and prevents fewer attacks overall.

\begin{figure}[t]
\centering
\includegraphics[width=1\columnwidth]{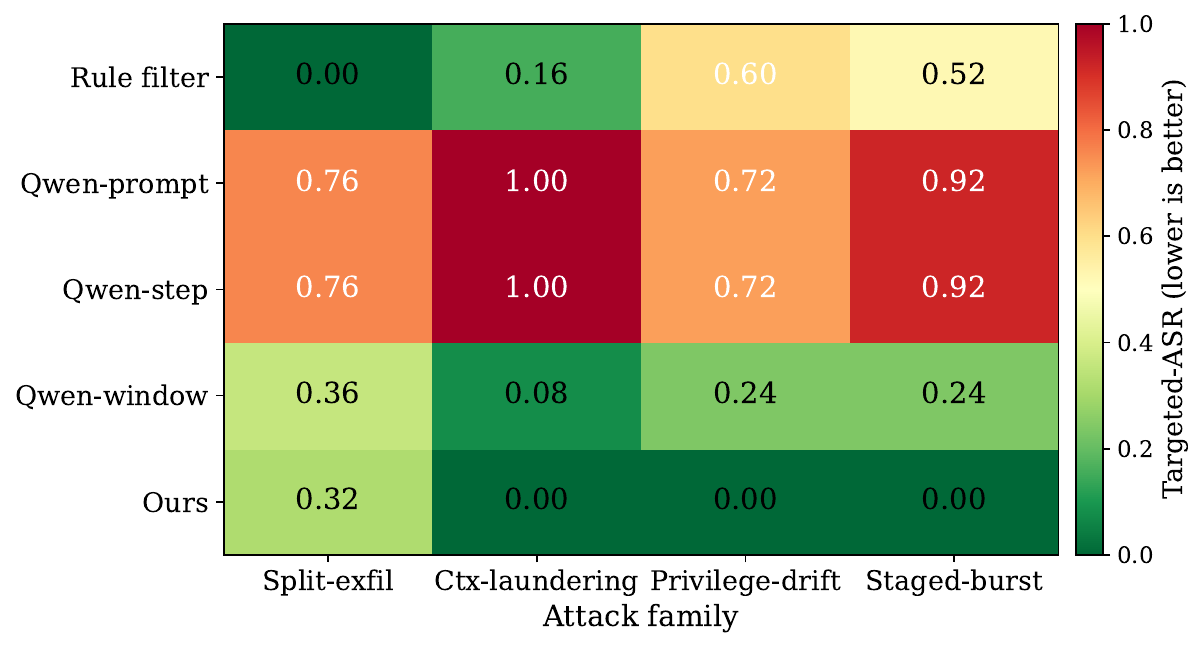}
\caption{Targeted attack-success rate by detector and attack family.}
\label{fig:per_family}
\end{figure}

Second, the latency gap is significant. The three Qwen-based baselines cost approximately $43$~ms of forward-pass time per prefix on an A6000 GPU, consuming more than $5\%$ of the LLM-planner latency budget and missing the $1\%$ deployability target marked in Figure~\ref{fig:frontier}. Our XGBoost detector runs in $4.6$~ms on a
single CPU thread with $0.6\%$ of the budget, placing it in the
upper-left corner of the latency-versus-$F_1$ plane,
$9.4$--$9.5{\times}$ faster than any LLM baseline while achieving the highest $F_1$ in the table.
Moreover,
Figure~\ref{fig:per_family} shows the targeted attack-success rate
broken down by family. The \textit{Rule-filter} catches sensitive filenames
visible in a single turn but largely misses multi-turn families. The
single-turn Qwen baselines fail uniformly,
confirming that single-turn embeddings, however rich, cannot recover the trajectory signal.

\begin{table}[t]
\centering
\setlength{\tabcolsep}{4pt}
\begin{tabular}{l l r r r r}
\toprule
Mode & Group(s) & $|\mathbf{z}|$ & AUC & $F_1$ & ASR\,red. \\
\midrule
\multirow{5}{*}{Isolated}
 & prompt   & 11 & $0.81$ & $0.57$ & $0.62$ \\
 & session  &  8 & $0.86$ & $0.74$ & $0.86$ \\
 & tool     &  6 & $0.65$ & $0.00$ & $0.00$ \\
 & context  &  6 & $0.67$ & $0.27$ & $0.33$ \\
\rowcolor{gray!12}
 & \textbf{trajectory} & 11 & $\mathbf{0.91}$ & $\mathbf{0.76}$ & $\mathbf{0.87}$ \\
\midrule
\multirow{5}{*}{\shortstack[l]{Leave-\\one-out}}
 & $-$prompt    & 31 & $0.92$ & $0.75$ & $0.87$ \\
 & $-$session   & 34 & $0.96$ & $0.65$ & $0.86$ \\
 & $-$tool      & 36 & $0.96$ & $0.77$ & $0.95$ \\
 & $-$context   & 36 & $0.96$ & $0.82$ & $0.97$ \\
\rowcolor{gray!12}
 & $\mathbf{-\textbf{trajectory}}$ & 31 & $\mathbf{0.92}$ & $\mathbf{0.68}$ & $\mathbf{0.76}$ \\
\midrule
Full & all five groups & 42 & $0.96$ & $0.81$ & $0.94$ \\
\bottomrule
\end{tabular}
\caption{Feature ablation. Each row trains our detector on
the indicated feature subset; \textbf{ASR\,red.} is the fraction of attack interactions blocked by the calibrated policy.}
\label{tab:ablation}
\end{table}

\begin{figure*}[t]
\centering
\includegraphics[width=1.6\columnwidth]{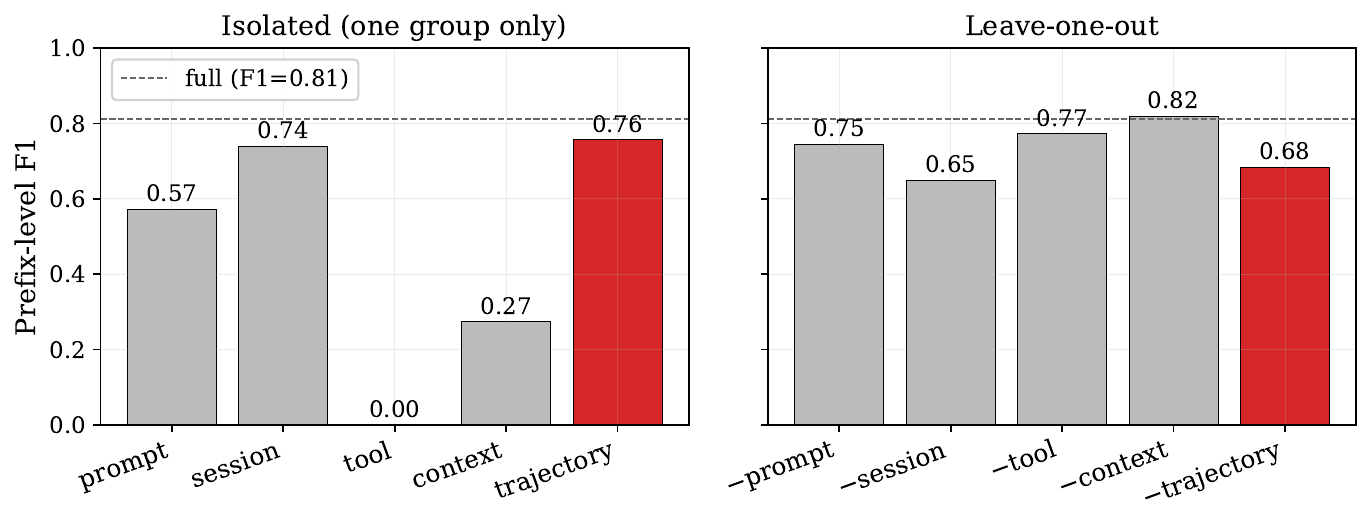}
\caption{Feature-group ablation. \emph{Left:} isolated; trajectory alone nearly
matches the full detector. \emph{Right:} leave-one-out; removing trajectory
 causes the largest drop. Dashed line marks full-model $F_1$.}
 \label{fig:ablation}
 \end{figure*}
\subsection{Ablation Study}
\label{sec:ablation}

We examine which feature groups drive detection performance by
decomposing the $42$-feature space into its five constituent groups:
prompt, session, tool, context, and trajectory. We train two
ablation variants: an \emph{isolated} model using only a single feature
group, and a \emph{leave-one-out} model that removes one group from the
full set. Results are reported in Table~\ref{tab:ablation} and
visualized in Figure~\ref{fig:ablation}.

The two analyses yield a consistent conclusion. In isolation, the
trajectory feature group achieves $F_1 = 0.76$ and an ASR reduction
of $0.87$, nearly matching the full model ($0.81$ and $0.94$
respectively) and substantially outperforming every other individual
group. Session features provide secondary lift ($F_1 = 0.74$), while prompt, tool, and context features in isolation are insufficient on their own. The leave-one-out analysis confirms this finding from the opposite direction. Removing the trajectory group causes the largest degradation: $F_1$ drops from $0.81$ to $0.68$ and ASR reduction falls
from $0.94$ to $0.76$, admitting an extra $18\%$ of attack interactions.
Removing any other individual group has a neutral or marginal effect,
suggesting that the prompt, tool, and context groups are largely
redundant with the trajectory features for this task.

%% file: sections/5_discussion.tex
\subsection{Discussion}

The results support the central claim of the paper: adversarial interaction is a broader and more realistic target than isolated prompt injection. Prompt-level defenses remain useful, but they do not adequately capture multi-step behavior, indirect content attacks, or gradual escalation. By modeling structured interaction signals over time, the proposed framework adds a complementary line of defense that is especially well suited to action-taking agents.

Our detector framework has practical strengths. It is lightweight, deployable, and compatible with existing guardrails. Its features are interpretable enough to support monitoring and debugging, and it does not require another LLM in the critical path. These properties make it attractive for systems where latency, cost, and operational transparency matter.

Several extensions of the framework are worth exploring. First, the features used here are structure data inspired by fraud detection domain and low latency; future work could investigate whether learned representations of session or trajectory dynamics can improve performance as well as robustness. Second, the framework currently assumes a static feature set; online adaptation to emerging attack patterns through continual learning would improve resilience against adaptive adversaries. Third, as standardized multi-turn agent security benchmarks become available, systematic evaluation across agent frameworks and tool sets will be needed to characterize generalization.



%% file: sections/6_conclusion.tex
\section{Conclusion}

In this work, we propose a new perspective on securing LLM agents by focusing on adversarial interaction patterns rather than isolated prompt classification. Prompt injection remains an important security issue, but is best understood as a special case within a larger class of risky interaction trajectories. To address this problem, we introduce a low-latency fraud detection layer that models structured signals from prompts, sessions, tool usage, context and cross-turn fraud signals.

The resulting framework preserves the deployment advantages of low-latency lightweight models while extending agent security beyond prompt-only defense. It supports real-time operation, earlier intervention, and better alignment with the security logic used in mature fraud detection systems. More broadly, this work suggests that interaction-level behavioral modeling should become an important component of deployment-time defense for LLM agents.
